# A New Algorithm Based Entropic Threshold for Edge Detection in Images

Mohamed A. El-Sayed

CS dept, Faculty of Computers and Information Systems , Taif Univesity, 21974 Taif, KSA
Mathematics department, Faculty of Science, Fayoum University, 63514 Fayoum, Egypt

**Abstract**
Edge detection is one of the most critical tasks in automatic image analysis. There exists no universal edge detection method which works well under all conditions. This paper shows the new approach based on the one of the most efficient techniques for edge detection, which is entropy-based thresholding. The main advantages of the proposed method are its robustness and its flexibility. We present experimental results for this method, and compare results of the algorithm against several leading edge detection methods, such as Canny, LOG, and Sobel. Experimental results demonstrate that the proposed method achieves better result than some classic methods and the quality of the edge detector of the output images is robust and decrease the computation time.

## 1. Introduction

Edge detection is an important field in image processing. It can be used in many applications such as segmentation, registration, feature extraction, and identification of objects in a scene. An effective edge detector reduces a large amount of data but still keeps most of the important feature of the image. Edge detection refers to the process of locating sharp discontinuities in an image. These discontinuities originate from different scene features such as discontinuities in depth, discontinuities in surface orientation, and changes in material properties and variations in scene illumination. [6]

Many operators have been introduced in the literature, for example Roberts, Sobel and Prewitt [2, 3, 5 , 8, 12, 15]. Edges are mostly detected using either the first derivatives, called gradient, or the second derivatives, called Laplacien. Laplacien is more sensitive to noise since it uses more information because of the nature of the second derivatives.

Most of the classical methods for edge detection based on the derivative of the pixels of the original image are Gradient operators, Laplacian and Laplacian of Gaussian (LOG) operators [14].

Gradient based edge detection methods, such as Roberts, Sobel and Prewitts, have used two 2-D linear filters to process vertical edges and horizontal edges separately to approximate first-order derivative of pixel values of the image. Marr and Hildreth achieved this by using the Laplacian of a Gaussian (LOG) function as a filter [10]. The paper [11] classified and comparative studies of edge detection algorithms are presented. Experimental results prove that Boie-Cox, Shen- Castan and Canny operators are better than Laplacian of Gaussian (LOG), while LOG is better than Prewitt and Sobel in case of noisy image.

The paper [6] used 2-D gamma distribution, the experiment showed that the proposed method obtained very good results but with a big time complexity due to the big number of constructed masks.

To solve these problems, the study proposed a novel approach based on information theory, which is entropy-based thresholding. The proposed method is decrease the computation time. The results were very good compared with the well-known Sobel gradient [7] and Canny [4] gradient results.

This paper is organized as follows: in Section 2 presents some fundamental concepts of the mathematical setting of the threshold selection. Section 3, we describe the proposed method of edge detection, and we describe the proposed algorithm. In Section 4, we report the effectiveness of our method when applied to some real-world and synthetic images, and compare results of the algorithm against several leading edge detection methods, such as Canny, LOG, and Sobel  method. At last conclusion of this paper will be drawn in Section 5.

## 2. Image thresholding

The set of all source symbol probabilities is denoted by $P$, $P= \{p_1, p_2, p_3, ..., p_k \}$. This set of probabilities must satisfy the condition $\sum_{i=1}^{k} p_i = 1$, $0\leq p_i \leq 1$. The average information per source output, denoted $S(Z)$ [9], Shannon entropy may be described as:

$$S(Z) = -\sum_{i=1}^{k} p_i \, ln(p_i) \quad (1)$$

being $k$ the total number of states. If we consider that a system can be decomposed in two statistical independent subsystems $A$ and $B$, the Shannon entropy has the extensive property (additivity) $S(A+B) = S(A) + S(B)$., this formalism has been shown to be restricted to the Boltzmann-Gibbs-Shannon (BGS) statistics.

However, for non-extensive systems, some kind of extension appears to become necessary. Tsallis [13] has proposed a generalization of the BGS statistics which is useful for describing the thermo statistical properties of non-extensive systems. It is based on a generalized entropic form,

$$S_q = \frac{1}{q-1}(1 - \sum_{i=1}^{k} p_i^q) \quad (2)$$

where the real number $q$ is a entropic index that characterizes the degree of non-extensivity. This expression recovers to BGS entropy in the limit $q \to 1$. Tsallis entropy has a non-extensive property for statistical independent systems, defined by the following rule [1]:

$$S_q(A+B) = S_q(A) + S_q(B) + (1-q) \cdot S_q(A) \cdot S_q(B). \quad (3)$$

Similarities between Boltzmann-Gibbs and Shannon entropy forms give a basis for possibility of generalization of the Shannon's entropy to the Information Theory. This generalization can be extended to image processing areas, specifically for the image segmentation, applying Tsallis entropy to threshold images, which have non-additive information content.

Let $f(x, y)$ be the gray value of the pixel located at the point $(x, y)$. In a digital image $\{f(x,y) | x \in \{1,2,...,M\}, y \in \{1,2,...,N\}\}$ of size $M \times N$, let the histogram be $h(a)$ for $a \in \{0,1,2,...,255\}$ with $f$ as the amplitude (brightness) of the image at the real coordinate position $(x, y)$. For the sake of convenience, we denote the set of all gray levels $\{0,1,2,...,255\}$ as $G$. Global threshold selection methods usually use the gray level histogram of the image. The optimal threshold $t^*$ is determined by optimizing a suitable criterion function obtained from the gray level distribution of the image and some other features of the image.

Let $t$ be a threshold value and $B = \{b_0, b_1\}$ be a pair of binary gray levels with $\{b_0, b_1\} \in G$. Typically $b_0$ and $b_1$ are taken to be 0 and 1, respectively. The result of thresholding an image function $f(x, y)$ at gray level $t$ is a binary function $f_t(x, y)$ such that $f_t(x,y) = b_0$ if $f_t(x,y) \le t$ otherwise, $f_t(x,y) = b_1$. In general, a thresholding method determines the value $t^*$ of $t$ based on a certain criterion function. If $t^*$ is determined solely from the gray level of each pixel, the thresholding method is point dependent [9].

Let $p_i = p_1, p_2, \ldots, p_k$ be the probability distribution for an image with $k$ gray-levels. From this distribution, we derive two probability distributions, one for the object (class $A$) and the other for the background (class $B$), given by:

$$\begin{aligned} p_A &: \frac{p_1}{P_A}, \frac{p_2}{P_A}, \ldots, \frac{p_t}{P_A}, \\ p_B &: \frac{p_{t+1}}{P_B}, \frac{p_{t+2}}{P_B}, \ldots, \frac{p_k}{P_B} \end{aligned} \quad (4)$$

and where

$$P_A = \sum_{i=1}^{t} p_i \quad, \quad P_B = \sum_{i=t+1}^{k} p_i \quad (5)$$

The Tsallis entropy of order $q$ for each distribution is defined as:

$$\begin{aligned} S_q^A(t) &= \frac{1}{q-1}(1 - \sum_{i=1}^{t} p_A^q), \\ \text{and} \quad S_q^B(t) &= \frac{1}{q-1}(1 - \sum_{i=t+1}^{k} p_B^q) \end{aligned} \quad (6)$$

The Tsallis entropy $S_q(t)$ is parametrically dependent upon the threshold value $t$ for the foreground and background. It is formulated as the sum each entropy, allowing the pseudo-additive property, defined in equation (2). We try to maximize the information measure between the two classes (object and background). When $S_q(t)$ is maximized, the luminance level $t$ that maximizes the function is considered to be the optimum threshold value [4].

$$t^*(q) = \underset{t \in G}{Arg \, max}[S_q^A(t) + S_q^B(t) + (1-q).S_q^A(t).S_q^B(t)]. \quad (7)$$

In the proposed scheme, first create a binary image by choosing a suitable threshold value using Tsallis entropy. The technique consists of treating each pixel of the original image and creating a new image, such that $f_t(x,y) = 0$ if $f_t(x,y) \le t^*(q)$ otherwise, $f_t(x,y) = 1$ for every $x \in \{1,2,...,M\}$, $y \in \{1,2,...,N\}$.

When $q \to 1$, the threshold value in Equation (2), equals to the same value found by Shannon's method. Thus this proposed method includes Shannon's method as a special case. The following expression can be used as a criterion function to obtain the optimal threshold at $q \to 1$.

$$t^*(1) = \underset{t \in G}{Arg\ max}[S^A(t) + S^B(t)]. \qquad (8)$$

The *Threshold* procedure to select suitable threshold value $t^*$ and $q$ can now be described as follows:

**Procedure *Threshold*,**
**Input:** A digital grayscale image *A* of size $M \times N$.
**Output:** The suitable threshold value $t^*$ of *A*, for $q \geq 0$.
Begin
  1. Let $f(x, y)$ be the original gray value of the pixel at the point $(x, y)$, $x=1..M$, $y=1..N$.
  2. Calculate the probability distribution $0 \leq p_i \leq 255$.
  3. For all $t \in \{0,1,…,255\}$,
      i. Apply Equations (4 and 5) to calculate $P_A$, $P_B$, $p_A$ and $p_B$.
      ii. Apply Equation (7) to calculate optimum threshold value $t^*$.
End.

## 3. The edge detection

We will use the usual masks for detecting the edges. A spatial filter mask may be defined as a matrix *w* of size $m \times n$. Assume that $m=2\alpha+1$ and $n=2\beta+1$, where α, β are nonzero positive integers. For this purpose, smallest meaningful size of the mask is 3×3. Such mask coefficients, showing coordinate arrangement as Figure 1.a.

| w(-1,-1) | w(-1,0) | w(-1,1) |
|---|---|---|
| w(0,-1) | w(0,0) | w(0,1) |
| w(1,-1) | w(1,0) | w(1,1) |

Fig 1.a

| f(x-1, y-1) | f(x-1,y) | f(x-1, y+1) |
|---|---|---|
| f(x, y-1) | f(x, y) | f(x, y+1) |
| f(x+1, y-1) | f(x+1,y) | f(x+1, y+1) |

Fig. 1.b

| 1 | 1 | 1 |
|---|---|---|
| 1 | × | 1 |
| 1 | 1 | 1 |

Fig. 1.c

Image region under the above mask is shown as Figure 1.b. In order to edge detection, firstly classification of all pixels that satisfy the criterion of homogeneousness, and detection of all pixels on the borders between different homogeneous areas. In the proposed scheme, first create a binary image by choosing a suitable threshold value using Tsallis entropy. Window is applied on the binary image. Set all window coefficients equal to 1 except centre, centre equal to × as shown in Figure 1.c.

Move the window on the whole binary image and find the probability of each central pixel of image under the window. Then, the entropy of each central pixel of image under the window is calculated as $S(CPix) = -p_c\ ln(p_c)$.

Table 1: p and S of central under window

| P | S | P | S |
|---|---|---|---|
| 1/9 | 0.2441 | 6/9 | 0.2703 |
| 2/9 | 0.3342 | 7/9 | 0.1955 |
| 3/9 | 0.3662 | 8/9 | 0.1047 |
| 4/9 | 0.3604 | 9/9 | 0.0 |
| 5/9 | 0.3265 | - | - |

where, $p_c$ is the probability of central pixel *CPix* of binary image under the window. When the probability of central pixel, $p_c = 1$, then the entropy of this pixel is zero. Thus, if the gray level of all pixels under the window homogeneous, $p_c = 1$ and $S = 0$. In this case, the central pixel is not an edge pixel. Other possibilities of entropy of central pixel under window are shown in Table 1.

In cases $p_c = 8/9$, and $p_c = 7/9$, the diversity for gray level of pixels under the window is low. So, in these cases, central pixel is not an edge pixel. In remaining cases, $p_c \leq 6/9$, the diversity for gray level of pixels under the window is high.

The complete *EdgeDetection* **Procedure** can now be described as follows:

**Procedure *EdgeDetection*;**
**Input:** A grayscale image *A* of size $M \times N$ and $t^*$.
**Output:** The edge detection image *g* of *A*.
Begin
  **Step 1**: Create a binary image: For all *x*, *y*,
      If $f(x, y) \leq t^*$ then $f(x, y) = 0$ Else $f(x, y) = 1$.
  **Step 2:** Create a mask, *w*, with 3×3, $a = (m-1)/2$ and $b = (n-1)/2$. (see Figure 1)
  **Step 3:** Create an M×N output image, g: For all x and y, Set $g(x, y) = f(x, y)$.
  **Step 4:** Checking for edge pixels:
      For all $y \in \{b+1, …, N-b\}$, and $x \in \{a+1, …, M-a\}$,
        sum = 0;
        For all $k \in \{-b, …, b\}$, and $j \in \{-a, …, a\}$,
          If ($f(x, y) = f(x+j, y+k)$) Then sum = sum+1.
        If (sum > 6) Then $g(x,y)=0$ Else $g(x,y)=1$.

(see Table 1)
End Procedure.

The steps of proposed algorithm are as follows:
Step1: We use Shannon entropy, the equation (8), to find the global threshold value ($t_1$). The image is segmented by $t_1$ into two parts, the object and the background. See Figure 2.

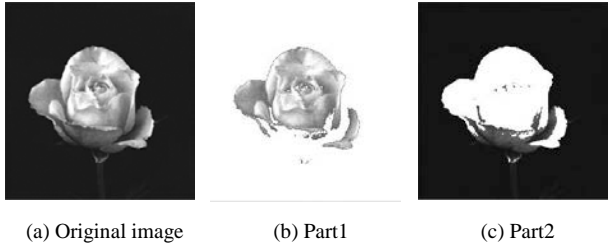

(a) Original image     (b) Part1     (c) Part2

Fig. 2. Original image , and its parts, Part1 and Part2.

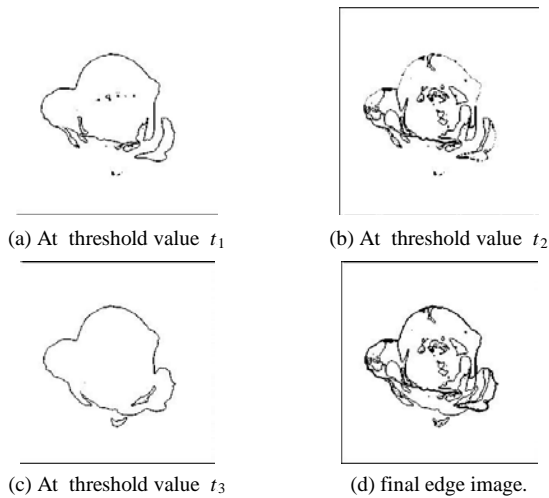

(a) At threshold value $t_1$     (b) At threshold value $t_2$

(c) At threshold value $t_3$     (d) final edge image.

Fig. 3 Edge images of original image , its parts, Part1 and Part2 and final output of edge image

Step2: We use Tsallis entropy, the equation (7) , $q=0.5$, Since, we can write the Equation (6) as:

$$S_{0.5}^A(t) = 2\sum_{i=1}^{t}\left|\sqrt{p_A}\right| - 2 ,$$

$$\text{and } S_{0.5}^B(t) = 2\sum_{i=t+1}^{k}\left|\sqrt{p_B}\right| - 2 \quad (9)$$

Therefore, we have

$$t^*(0.5) = Arg\max_{t\in G}[(\sum_{i=1}^{t}\left|\sqrt{p_A}\right|)(\sum_{i=t+1}^{k}\left|\sqrt{p_B}\right|) - 1]. \quad (10)$$

Applying the equation (10), to find the locals threshold values ($t_2$) and ($t_3$) of Part1 and Part2, respectively.

Step3: Applying *EdgeDetection* Procedure with threshold values $t_1$, $t_2$ and $t_3$. See Figure 3 .a-c

Step4: Merge the resultant images of step 3 in final output edge image. See Figure 3.d

In order to reduce the run time of the proposed algorithm, we make the following steps:
Firstly, the run time of arithmetic operations is very much on the M×N big digital image, $I$ , and its two separated regions, Part1 and Part2. We are use the linear array $p$ (probability distribution) rather than $I$ , for segmentation operation, and threshold values computation $t_1$, $t_2$ and $t_3$. Secondly, rather than we are create many binary matrices $f$ and apply the edge detector procedure for each region individually, then merge the resultant images into one. We are create one binary matrix $f$ according to threshold values $t_1$, $t_2$ and $t_3$ together, then apply the edge detector procedure one time. This modifications will reduce the run time of computations.

The following procedures summarize the proposed technique for calculating the optimal threshold values and the edge detector. The above procedures can be done together in the following MATLAB program:

*MainProgram.m file*

```
I=imread('Lena.tif');
[M,N]=size(I);
p = zeros(256,3);
for ii=1:256 p(ii,1)=ii-1; end;
p(:,2) = imhist(I);
p (p(:,2)==0,:) = []; % remove zero entries in p
% Calling Shannon procedure, return t1 value and
its location in p
[T1,Loc]=Shannon(p);
% Calling Tsallis procedure of Part1
pLow= p(1:Loc,:);       T2= Tsallis_Sqrt(pLow);
% Calling Tsallis procedure of Part2
pHigh=p(Loc+1:size(p),:); T3=Tsallis_Sqrt(pHigh);
% Cerate binary matrices  f
f=zeros(M,N);
for i=1:M;
  for j=1:N;
    if ((I(i,j)>= T2)&(I(i,j)<T1))|(I(i,j)>= T3)
        f(i,j)=1; end;
  end;
end
% Calling EdgeDetector procedure, return edge
detection image.
[g]= EdgeDetector(f);
figure;
imshow(g);
```

*Shannon.m file*

```
function [T,Loc]=Shannon(p)
p(:,3) = p(:,2)./ sum(p(:,2)); % normalize p so
that sum(p) is one.
[M1, N1]=size(p); Max1= 0;
for t=1 : size(p)
    PA= sum(p(1:t,3));
    PB= 1-PA ;
    p1=p(1:t,3)./PA; % p1 is i probability in PA
    p2=p(t+1:M1,3)./PB; %p2 is i prob. in PB
    Sa= -sum(p1.*log2(p1));
    Sb= -sum(p2.*log2(p2));
```

```
    Sab= Sa + Sb;
    if(Sab>Max1)  T=p(t,1); Loc=t;Max1=Sab; end;
end;
```

*Tsallis_Sqrt.m file*

```
function [T]=Tsallis_Sqrt(p)
p(:,3) = p(:,2)./ sum(p(:,2));
[M1, N1]=size(p); Max1= 0;
for t=1 : M1
    PA= sum(p(1:t,3));
    PB= 1-PA ;
    p1= p(1:t ,3)./PA;
    p2= p(t+1 :M1 ,3)./PB;
    Tab=sum(sqrt(p1))*sum (sqrt(p2))-1;
    if ( Tab > Max1 ) T=p(t,1); Max1=Tab; end;
end;
```

*EdgeDetector.m file*

```
function [g]=EdgeDetector(f);
[M,N]=size(f);
g=zeros(M,N);
for y=2 : N-1;
  for x=2 : M-1;
    sum1=0;
    for k=-1:1;
        for j=-1:1;
           if(f(x,y)==f(x+j,y+k))
              sum1=sum1+1;
           end;
        end;
    end;
    if (sum1<=6)  g(x,y)=1;  end;
  end;
end;
```

## 4. Results and Discussions

In order to test the method proposed in this paper and compare with the other edge detectors, common gray level test images with different resolutions and sizes are detected by Canny, LOG, and Sobel and the proposed method respectively. The performance of the proposed scheme is evaluated through the simulation results using MATLAB. Prior to the application of this algorithm, no pre-processing was done on the tested images.

As the algorithm has two main phases – global and local enhancement phase of the threshold values and detection phase, we present the results of implementation on these images separately. Here, we have used in addition to the original gray level function $f(x, y)$, a function $g(x, y)$ that is the average gray level value in a $3\times 3$ neighborhood around the pixel $(x, y)$.

The proposed algorithm used the good characters of each Shannon entropy and Tsallis entropy, together, to calculate the global and local threshold values. Hence, we ensure that the proposed algorithm done better than the algorithms that based on Shannon entropy or Tsallis entropy separately.

Though the performance of the proposed entropic edge detector excels as a shape and detail detector, it is fraught with some drawbacks. It fails to provide all thinned edges. The presence of thick edges at some locations needs to be addressed by the proper choice of parameter $q$. The weak edges are not eliminated but for some applications, these may be required. This detector has another distinctive feature, i.e. it retains the texture of the original image. This feature can be utilized for the identification of fingerprints, where the ridges may have different intensities. As the success of the edge detection depends on these parameters, we are experimenting on several images to come up with a useful selection guideline with $0<q<1$.

We run the Canny, LOG, and Sobel methods and the proposed algorithm 10 times for each image with different sizes. As shown in Figures 4-7, The charts of the test images and the average of run time for the classical methods and proposed scheme. It has been observed that the proposed edge detector works effectively for different gray scale digital images as compare to the run time of Canny method.

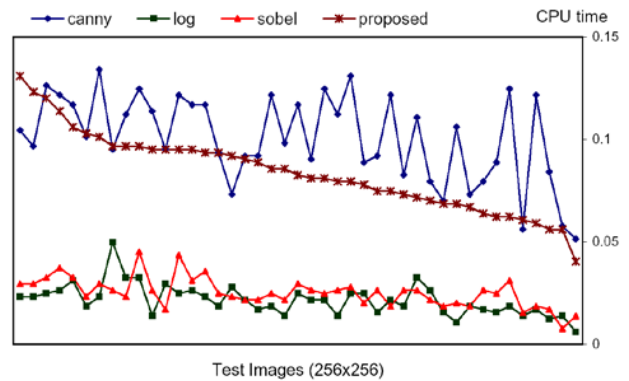

Fig. 4. CPU time of Canny, LOG, Sobel, and proposed method with 256×256 pixel test images

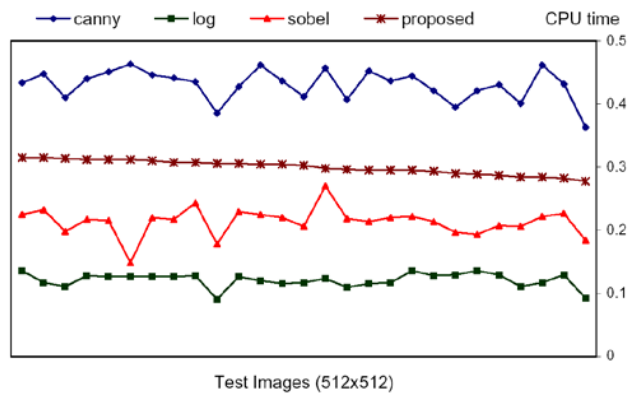

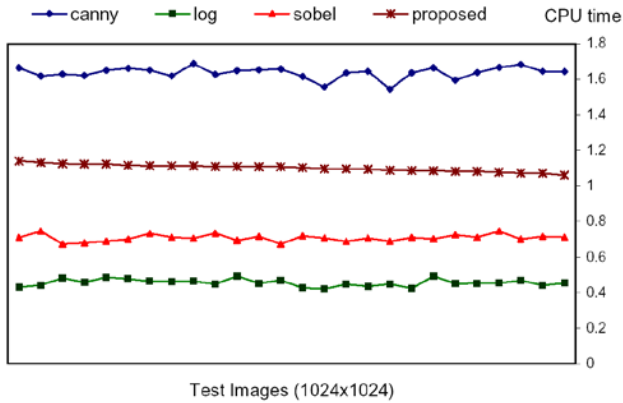

Fig. 5. CPU time of Canny, LOG, Sobel, and proposed method with 512×512 pixel test images

Fig. 6. CPU time of Canny, LOG, Sobel, and proposed method with 1024×1024 pixel test images

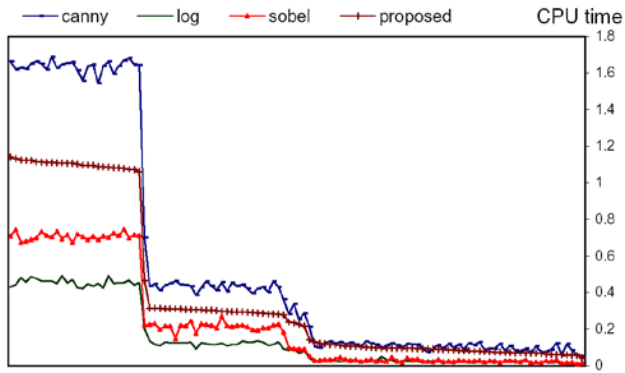

Fig. 7. CPU time of Canny, LOG, Sobel, and proposed method with different size test images

Some selected results of edge detections for these test images using the classical methods and proposed scheme are shown in Figures 8-12. From the results; it has again been observed that the proposed method works well as compare to the previous methods, LOG and Sobel (with default parameters in MATLAB).

## 5. Conclusion

The hybrid entropic edge detector presented in this paper uses both Shannon entropy and Tsallis entropy, together. It is already pointed out in the introduction that the traditional methods give rise to the exponential increment of computational time. However, the proposed method is decrease the computation time with generate high quality of edge detection. Experiment results have demonstrated that the proposed scheme for edge detection works satisfactorily for different gray level digital images. Another benefit comes from easy implementation of this method.

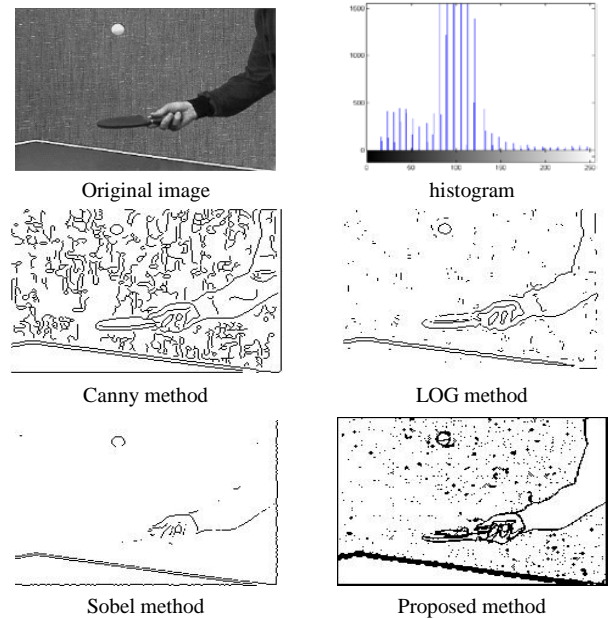

Fig. 8. Sport image with 224×153 pixel.

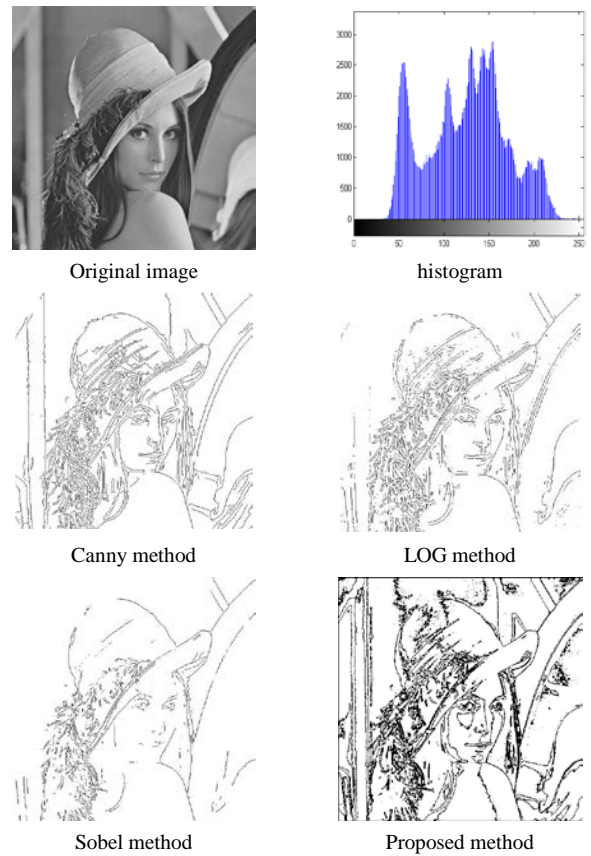

Fig. 9. Lena image with 512×512 pixel.

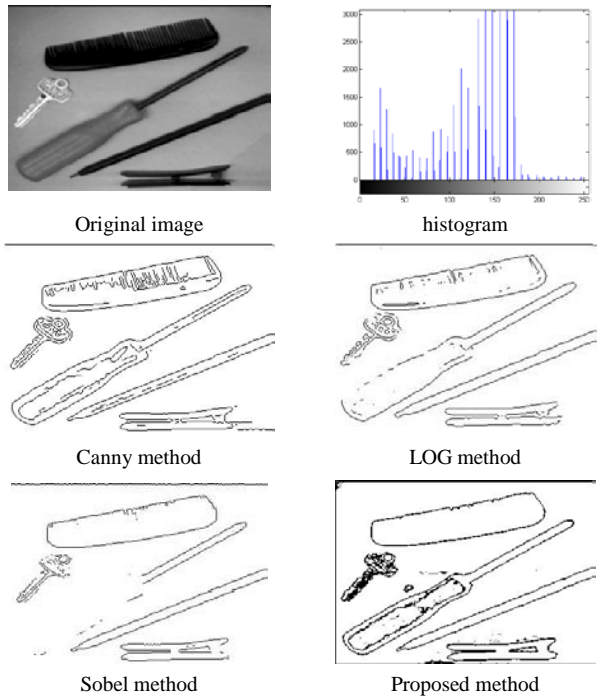

Fig. 10. Tools image with 322×228 pixel.

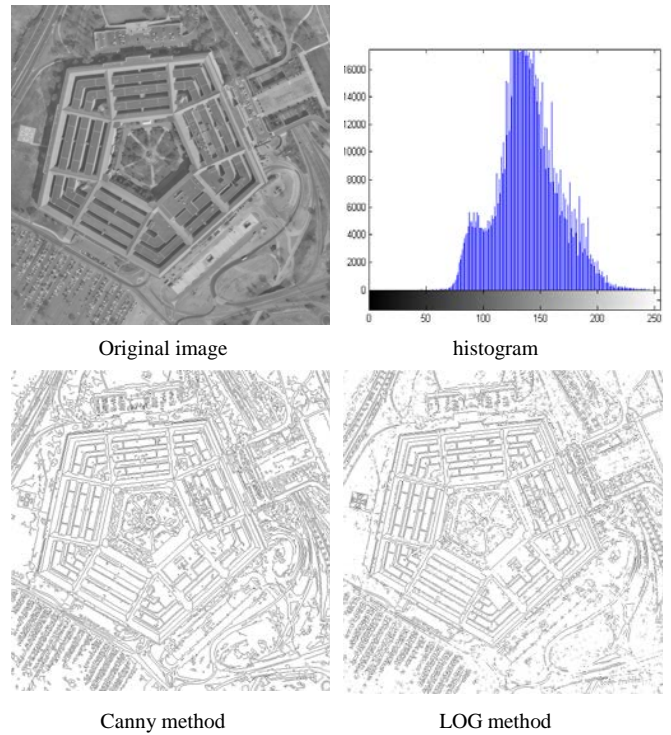

Fig. 12, pentagon image with 1024×1024 pixel.

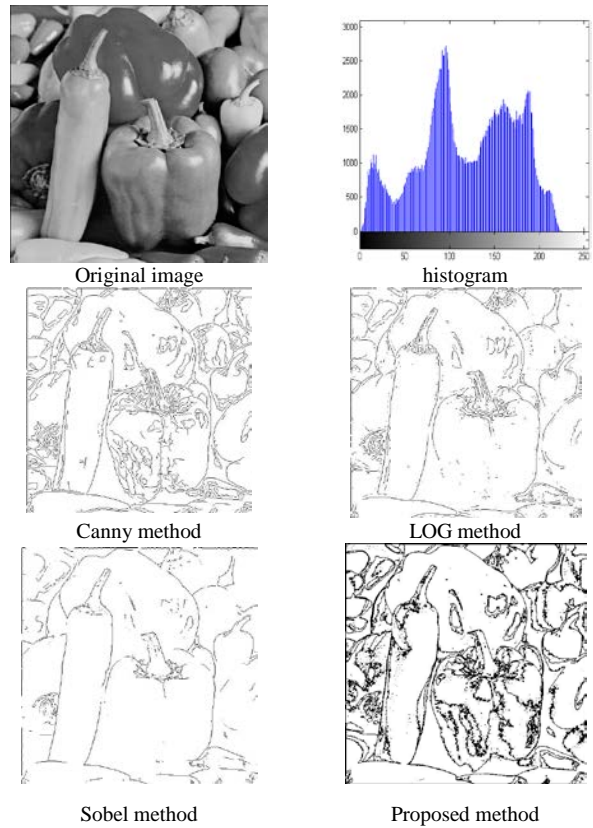

Fig. 11. Peppers image with 512×512 pixel.

**Biography**

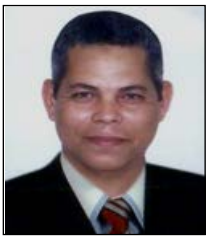

**Mohamed A. El-Sayed** received his B.Sc. from Faculty of Science (Maths & CS in June 1994) at Minia University - Egypt. His M.Sc. from Faculty of Science (Maths\CS) at South Valley University in 2002. Ph.D. degrees in Computer Science from Faculty of Science at Minia University in 2007. His research interests include image processing, computer graphic and graphs drawing. He has published several international journals and Conference papers in the above area. He is working in Mathematics department, Faculty of Science, Fayoum University, Egypt.  Currently, he is an Assistant professor of Computer Science at Faculty of Computers and Information Science , Taif Univesity,  KSA.